\documentclass[a4paper]{jpconf}

\usepackage{xspace}
\usepackage{amssymb}
\usepackage{amsmath}
\usepackage{graphicx}
\usepackage{float}
\usepackage{hyperref}
\usepackage[ruled,vlined]{algorithm2e}

\newcommand{\varpdf}{\ensuremath{\mathrm{Var}^{\mathrm{(pdf)}}}\xspace}
\newcommand{\varensemble}{\ensuremath{\mathrm{Var}^{\mathrm{(train)}}}\xspace}
\newcommand{\vartot}{\ensuremath{\mathrm{Var}^{\mathrm{(tot)}}}\xspace}
\newcommand{\yrone}{\ensuremath{Y_{r,1}}\xspace}
\newcommand{\yrtwo}{\ensuremath{Y_{r,2}}\xspace}
\newcommand{\yeone}{\ensuremath{Y_{e,1}}\xspace}
\newcommand{\yetwo}{\ensuremath{Y_{e,2}}\xspace}

\begin{document}
\title{Generative models uncertainty estimation}

\author{L Anderlini$^{1}$, C Chimpoesh$^{2}$, N Kazeev$^{2}$ and A Shishigina$^{2}$ on behalf of the LHCb collaboration}

\address{$^1$ Universita e INFN, Firenze (IT), via Sansone 1, 50019 Sesto Fiorentino (FI), Italia}
\address{$^2$ HSE University, 20 Myasnitskaya st., Moscow 101000, Russia}

\ead{nikita.kazeev@cern.ch}

\begin{abstract}
In recent years fully-parametric fast simulation methods based on generative models have been proposed for a variety of high-energy physics detectors. By their nature, the quality of data-driven models degrades in the regions of the phase space where the data are sparse. Since machine-learning models are hard to analyse from the physical principles, the commonly used testing procedures are performed in a data-driven way and can’t be reliably used in such regions. In our work we propose three methods to estimate the uncertainty of generative models inside and outside of the training phase space region, along with data-driven calibration techniques. A test of the proposed methods on the LHCb RICH fast simulation is also presented.
\end{abstract}


\section{Introduction}
In the last years, deep learning has become a common tool in natural science. Generative models, such as generative adversarial networks (GANs)~\cite{goodfellow2014generative}, variational autoencoder (VAE)~\cite{kingma2013auto}, normalising flows~\cite{rezende2015variational}, and diffusion models~\cite{sohl2015deep}, can learn to sample from a distribution efficiently. They are used for fully-parametric simulation of detectors -- in place of computationally-intensive simulation from the physical principles, usually with Geant4~\cite{DERKACH2020161804, Vallecorsa_2018, fedorGAN, AGOSTINELLI2003250}.

Neural networks (NN) are black-box models that don't provide theoretical guarantees on the uncertainty of the prediction. This makes it difficult to use them in rigorous scientific reasoning. Uncertainty of machine learning models is an active area of research, but almost all works deal with classification and regression tasks, not generative modelling~\cite{gawlikowski2022survey}. A recent work \cite{bellagente2021understanding} shows how Bayesian normalising flows capture uncertainties.

Our work extends uncertainty estimation research by introducing new methods for estimating the uncertainty of GANs. Comparing with \cite{bellagente2021understanding}, in practice GANs are usually faster in training and inference, and more accurate than normalising flows, and thus are more widely used for fast simulations in high-energy physics. The contributions of this work are summarised as follows:

\begin{itemize}
    \item We propose methods for estimating uncertainty of GANs
    \item We propose an approach to distillate the  ensemble into a single model for efficient uncertainty computation
\end{itemize}

\section{LHCb RICH fast simulation}
In the LHCb experiment, the new fully-parametric simulation of Ring-Imaging Cherenkov detectors (RICH) is based on training a fully-connected Cramer GAN~\cite{Maevskiy_2020, bellemare2017cramer} to approximate the reconstructed detector response. It is trained using the real data calibration samples~\cite{aaij2019selection}.

RICH particle identification works as follows. First, the likelihoods for each particle type hypothesis are computed for each track. Second, the delta log-likelihoods are computed as the difference between the given hypothesis and the pion hypothesis. The variables are named \texttt{RichDLL*}, where \texttt{*} can be \texttt{k} (kaon), \texttt{p} (proton), \texttt{mu} (muon), \texttt{e} (electron) and \texttt{bt} (below the threshold of emitting Cherenkov light).

For the generator, input $x \in X \subset  \mathbb{R}^{3 + d_\text{noise}}$ consists of kinematic characteristics of particles (pseudorapidity $\eta$, momentum $P$, number of tracks) and random noise. The output $y \in Y \subset \mathbb{R}^5$ corresponds to the delta log-likelihoods.

\section{Uncertainty estimation methods}
\subsection{MC dropout}
Common dropout \cite{JMLR:v15:srivastava14a} acts as a regularisation to avoid overfitting when training an NN. The dropout is applied at both training and inference for Monte Carlo dropout (MC dropout) \cite{gal2016dropout}. The prediction is no longer deterministic but depends on which NN nodes are randomly chosen to be kept. 
The MC dropout generates random predictions, and the latter has the interpretation of samples from a probabilistic distribution.

In our work, for MC dropout experiments, we add a dropout layer after each fully connected one and train with the same configuration as before. In the beginning, we used Bernoulli dropout, and then we experimented with Gaussian and Variational dropouts  \cite{molchanov2017variational}. Finally, we found that the ``structured'' dropout modification (neuron with the neighborhood of arbitrary size \textit{k} zeroed with probability \textbf{p}) improves uncertainty quality.

During inference, for each batch we generate a fixed set of dropout masks as a way to have a virtual ensemble.

\subsection{Adversarial deep ensembles}
Ensemble methods are a widely-used heuristic uncertainty estimation method~\cite{Lakshminarayanan_ens}. The core idea of ensembles is to introduce perturbations to the training procedure that shouldn't affect the outputs. Thus, the observed deviation in outputs is considered as uncertainty. 

These perturbations can be implemented using randomisation techniques such as bagging and random initialisation of the NN parameters. Bagging on average uses 63\% unique data points which leads to a biased performance estimate~\cite{Lakshminarayanan_ens}. The diversity of the ensemble also tends to zero with the increase of training dataset size. While a reasonable outcome for in-domain uncertainty, it renders bagging unsuitable for out-domain uncertainty estimation.

In our method we start with the idea of diversity through NN weights. In addition to random initialisation, we add a component to the loss function that rewards the models for being different. For Cramer GAN~\cite{bellemare2017cramer} the loss function is modified as follows:

\begin{equation}
f(y) = || D(y) - D(y_g') ||_2 - || D(y) ||_2,
\end{equation}
\begin{equation}
L_G = f(y_r) - f(y_g) - \pmb{\alpha || D(y_g) - D(y_{\bigcup g}) ||_2}, 
\label{eq:adversarial-loss}
\end{equation}
where $y_r$ are real data, $y_g$ are generated data, and $y_{\bigcup g}$ is a concatenation of the predictions of the ensemble,
corresponding to a model with averaged probability density; $\alpha \geq 0$ is a hyperparameter. The method is not specific to Cramer GAN and can be used with any GAN.

We reduce the influence of the adversarial component as the training progresses. Using pre-trained discriminators leads to more variety among the models. The overall training scheme is summarised in Algorithm \ref{alg:adversarial}. As $\alpha$ tends to zero, each ensemble member is trained without additional bias. This provide a principled advantage to adversarial ensembles: instead of heuristically perturbing the training objective as common in other methods~\cite{gawlikowski2022survey}, we take advantage of existence of many equally good local minima and search for a maximally diverse set of solutions to the unbiased problem of learning the distribution.

\begin{algorithm}[H]
\label{alg:adversarial}
\SetAlgoLined
1. Train several GANs with the classic loss ($\alpha = 0$)\;
2. Reinitialize the generators with random weights; keep the
discriminators weights; set $\alpha > 0$. We used $\alpha = 10$\;
3. Train both the generators and the discriminators with our loss Eq.~\eqref{eq:adversarial-loss},
decrease $\alpha$ gradually to $0$\;
 \caption{Adversarial ensembles training scheme}
\end{algorithm}

\subsection{Distillation}
Running multiple generators for each simulated particle is excessively expensive for a fast simulation. Methods for distilling ensemble models are discussed in the literature~\cite{malinin2019ensemble}, but they do not deal with generative models. 
Let us indicate the variance of the underlying PDF as \varpdf, and the variance due to differences among the trained generators, representing the uncertainty on the implicit model of the underlying PDF, as \varensemble.
Evaluating the ensemble multiple times, at fixed conditions, we expect a 
distribution of outputs with a variance $\vartot = \varpdf + \varensemble$ as long 
as we accept the very reasonable hypothesis that the random component 
of each generator in the ensemble is not correlated to the random differences
between generators of the ensemble. Let $Y$ be a random variable -- output of the generative model; \yrone and \yrtwo be the results of two independent inferences of a generator with fixed input conditions $X$ ($\eta$, $P$, number of tracks in the case of RICH GAN), then
\begin{equation}
    \varpdf(Y|X) = \frac{1}{2}\mathbb E_{\mathrm{reference}}\left[(Y_{r,2} - Y_{r,1})^2\right],
\end{equation}
where $\mathbb E_\mathrm{reference}$ indicates the average over several samples obtained from the same, reference model. Instead, when sampling \yeone and \yetwo from independent predictors in the
ensemble, we expect their variance to be 
\begin{equation}
    \vartot(Y|X) = \frac{1}{2}\mathbb E_{\mathrm{ensemble}}\left[(Y_{e,2} - Y_{e,1})^2\right]
\end{equation}

By training a regressor to predict  $(\yrone - \yrtwo)^2$ and $(\yeone - \yetwo)^2$ as a function of $X$ optimised to minimise the Mean Squared Error, we obtain explicit models for $2\cdot\varpdf(Y|X)$ and $2\cdot\vartot(Y|X)$, which can be combined to assess \varensemble as a function of
the conditions $X$ as 
\begin{equation}
    \varensemble(Y|X) = \vartot(Y|X) - \varpdf(Y|X)
\end{equation}

In summary, naming $f_r(X)$ and $f_e(X)$ the trained predictors for 
$(\yrone - \yrtwo)^2$ and $(\yeone - \yetwo)^2$, respectively,
and assuming a normal distribution for the training error, the 
uncertainty on the generated samples, at a given condition $X$, is estimated as 
\begin{equation}
    \sigma_{\mathrm{syst}}(X) = \sqrt{
        \frac{1}{2} f_e\left(X\right) - \frac{1}{2} f_r\left(X\right)}
\end{equation}

\section{Results}
\subsection{Figure of merit}
The quality of a fast simulation model in a particular phase space region is measured by the difference between the distributions of real and generated data. The objective of our uncertainty estimation methods is to predict this discrepancy.

To evaluate our methods, we compare background efficiency on real and generated data, computed as following. \texttt{RichDLL} values are commonly used for filtering tracks by a condition \texttt{RichDLLx > threshold}. We choose a threshold for \texttt{RichDLLx} so that 90\% of all tracks with type \texttt{x} in the training dataset are accepted. Since the detector is not perfect, not only \texttt{x}-particles pass the selection, but there are also false positives. We plot the efficiency of the requirement \texttt{RichDLLx > threshold} on pions, as an approximation of the background selection efficiency. For a good uncertainty estimate real efficiencies should lie inside the uncertainty bounds for 68\% of the bins.

\subsection{Uniform split}
\begin{figure}[!ht]
    \centering
	\includegraphics[width=0.7\textwidth]{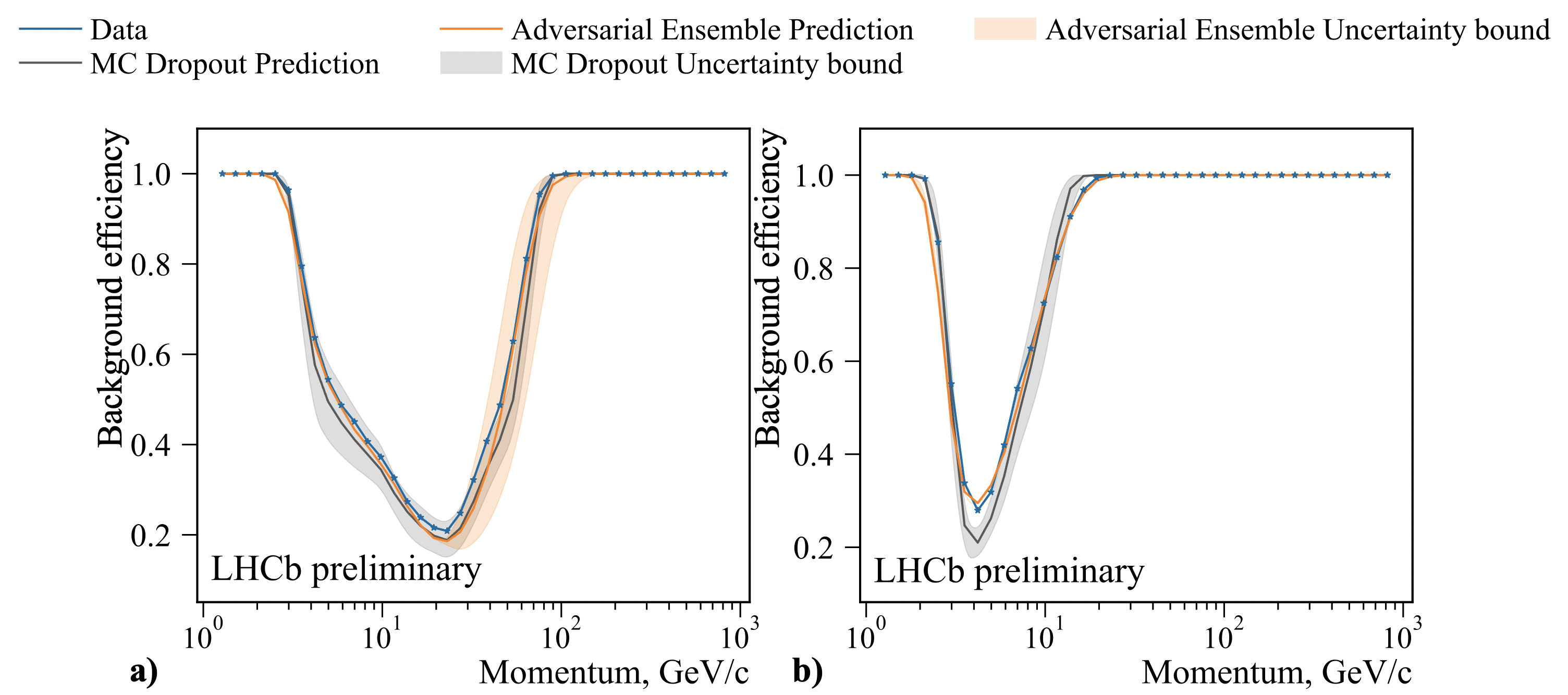}
	\hspace{0.3pc}%
    \begin{minipage}[b]{10.5pc}\caption{Fraction of pions in the test dataset selected by applying requirements on \texttt{RichDLLK} (a) and \texttt{RichDLLmu} (b) with thresholds corresponding to 90\% selection efficiency on kaons and muons, respectively.}
	\label{fig:uniform-split}
	\end{minipage}
\end{figure}

We uniformly split the dataset into training and testing parts, containing 2 and 1 million examples, correspondingly. The results are presented in figure \ref{fig:uniform-split}; for most of the bins efficiency on the test data lies inside the error bounds of the efficiency of the model. The side effect of MC dropout is dropout layers themselves. The dropout acts as a regularizer: it reduces variance but may increase the bias in the resulting model. Thus, the MC dropout model could be overregularized (more significant RichDLLmu efficiency error compared to ensembles).

\subsection{Extrapolation scan}
\begin{figure}[!ht]
    \centering
	\includegraphics[width=0.45\textwidth]{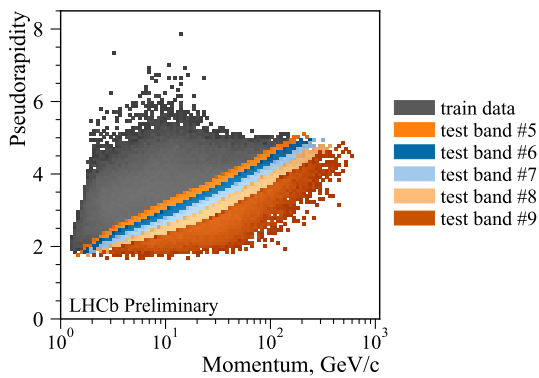}
	\hspace{2pc}%
    \begin{minipage}[b]{14pc}\caption{\label{fig:line-split}Training and testing datasets used for the extrapolation scan. The regions are separated by straight lines in the normalised space. Pions data correspond to the ones present in the LHCb Run 2 calibration sample~\cite{aaij2019selection}. Each test band contains the same number of examples.}
    \end{minipage}
\end{figure}

This test aims to assess the performance of the models in the regions of the phase space where there are no data. We emulate this situation by splitting the data into train and test parts in $P$ and $\eta$ space as shown in figure \ref{fig:line-split}.
The train part contains 947933 examples, and the test part contains 523917 examples. The models are trained on the train part. For evaluation, 101917 examples are samples from each training and test band. The high number of tracks makes the statistical uncertainty of the efficiency estimation negligible, both for real and generated data.

\begin{figure}[!ht]
    \centering
	\includegraphics[width=0.7\textwidth]{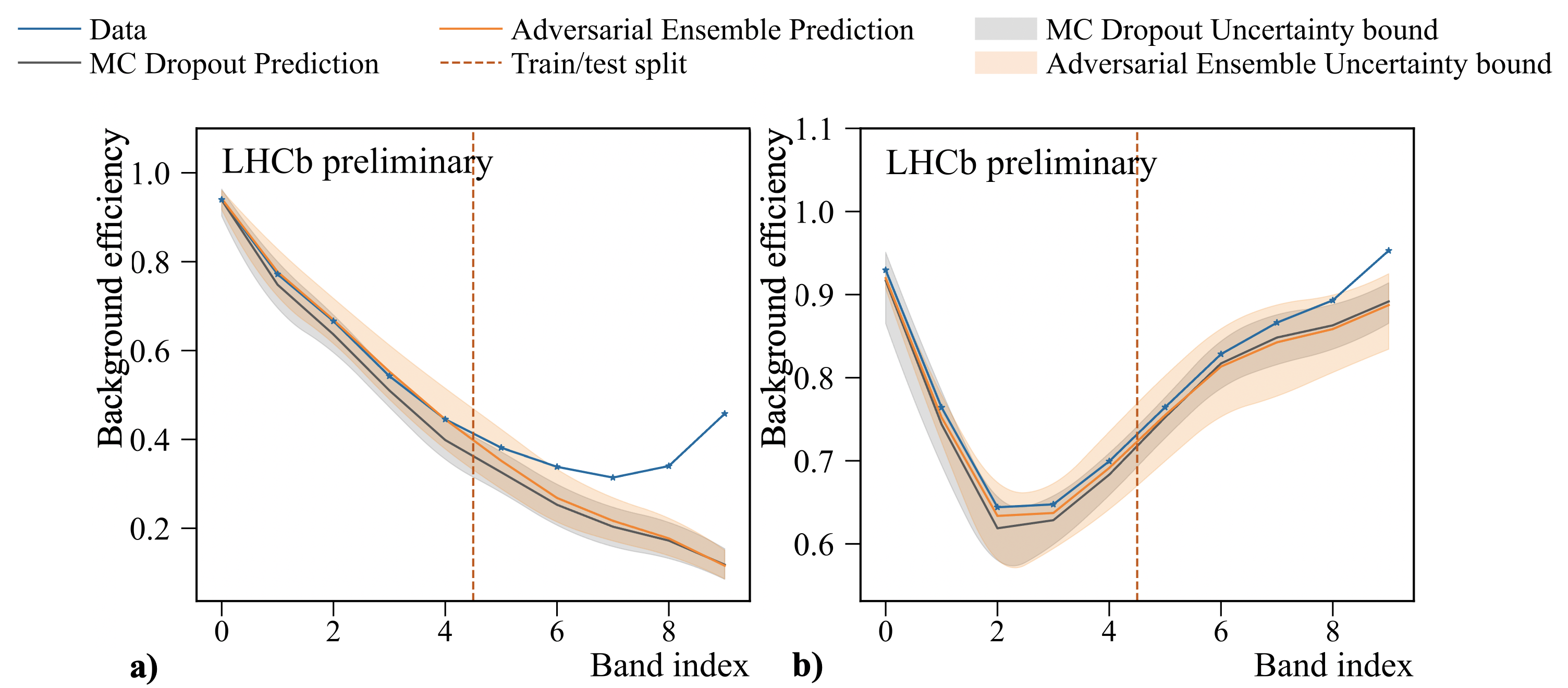}
	\hspace{0.3pc}%
    \begin{minipage}[b]{10.5pc}\caption{Fraction of pions selected by applying requirements on \texttt{RichDLLK} (a) and \texttt{RichDLLmu} (b) with thresholds corresponding to 90\% selection efficiency on kaons and muons, respectively as a function of the extrapolation scan test band index.}
	\label{fig:line-split-results}
	\end{minipage}
\end{figure}

The results are present in figure \ref{fig:line-split-results}. The adversarial ensembles show wider uncertainty bounds; nevertheless, both methods underestimate the uncertainty in the last bands. For kaons, figure \ref{fig:line-split-results}~(a), real background efficiency decreases for the first half of the test part, then starts to increase. This demonstrates a great obstacle for extrapolating with purely machine learning methods, as the qualitative change can't be reasonably predicted from the training data without incorporating additional knowledge. For muons,  figure~\ref{fig:line-split-results}~(b), efficiency in the test part continues the trend observed in the training part, resulting in uncertainty bands that contain the real efficiencies for the most bands.

\section{Conclusion}
We present methods for estimating uncertainty of GANs with adversarial ensembles and MC dropout. Although in this work we only use Cramer GAN, both methods are applicable to any GAN. The ensembles have a desirable theoretical property: each model converges to a local minimum of the unperturbed problem. We propose a method for distilling ensemble-based uncertainty estimation into a single model for fast inference.

The methods are evaluated on the LHCb RICH dataset. For most of the bins, efficiency on the test data lies inside the error bounds of the efficiency of the model. In the extrapolation case, the uncertainty increases while getting further from the training region. However, the uncertainty does not increase sufficiently to account for the discrepancy in the furthest test regions where the detector operational conditions are much different from those corresponding to the training sample. Our code is available online \cite{gan-uncertainty-mc, gan-uncertainty-ensembles}.

This work is a first step towards incorporating GAN uncertainty into high-energy physics fast simulation. We see the future directions for the research as following. Better correspondence between uncertainty and the real/simulated data difference could be achieved, along with more robust uncertainty growth outside the training region. We use background efficiency as a proxy metric, evaluating GAN uncertainty impact on the uncertainty of the final measurement would be more instructive.

\ack
We are grateful to Denis Derkach for sharing his knowledge of LHCb fast simulation and review. The publication was prepared within the framework of the Academic Fund Program at the HSE University in 2022 (grant №22-00-02). This research was supported in part through computational resources of HPC facilities at HSE University~\cite{Kostenetskiy_2021}.
\section*{References}
\bibliographystyle{iopart-num}
\bibliography{main}
\end{document}